\documentclass[lettersize, 10 pt, journal, twoside]{IEEEtran}
\usepackage{amsmath,amsfonts}
\usepackage{array}
\usepackage[caption=false,font=normalsize,labelfont=sf,textfont=sf]{subfig}
\usepackage{textcomp}
\usepackage{stfloats}
\usepackage{url}
\usepackage{verbatim}
\usepackage{graphicx}
\usepackage{cite}
\usepackage{lineno}
\usepackage{amssymb}
\usepackage{bm}
\usepackage{mathtools}
\usepackage{color, xcolor}
\usepackage[linesnumbered,ruled,vlined]{algorithm2e}

\usepackage{multirow}

\usepackage{subfig}

\usepackage{overpic}

\usepackage{booktabs}

\usepackage{amssymb}

\usepackage{bbding}

\usepackage{pifont}

\usepackage{threeparttable}

\usepackage{fancyhdr}

\definecolor{myred}{RGB}{255,78,0}    % 深红色
\definecolor{mygreen}{RGB}{34,139,34} 
\definecolor{myblue}{RGB}{0,0,128} 

\newcommand{\bred}[1]{\textcolor{myred}{\textbf{#1}}}
\newcommand{\bgreen}[1]{\textcolor{mygreen}{\textbf{#1}}}

\definecolor{myrevised}{RGB}{255,20,0}
\newcommand{\brevised}[1]{#1}

\begin{document}	
	
	\begin{titlepage}
		\centering
		\vspace*{5cm}
		\textbf{\Large IEEE Copyright Notice}
		\vspace{2cm}
		
		\begin{minipage}{0.8\textwidth}
			\large
			\copyright\ 2025 IEEE. 
			
			Personal use of this material is permitted. Permission from IEEE must be obtained for all other uses, in any current or future media, including reprinting/republishing this material for advertising or promotional purposes, creating new collective works, for resale or redistribution to servers or lists, or reuse of any copyrighted component of this work in other works.
			
			This file corresponds to the accepted version of the manuscript published in IEEE Robotics and Automation Letters, VOL. 11, NO. 1, pp. 282-289, JANUARY 2026. Digital
			Object Identifier 10.1109/LRA.2025.3632120
		\end{minipage}
		\vfill
	\end{titlepage}

%\title{A Direct Inertial Measurement Factor for Continuous-Time Trajectory Estimation on White-Noise-on-Jerk Prior}
%\title{Direct Inertial Factor on Differential $SE(3)$ for Continuous-Time Trajectory Estimation}
\title{Continuous Gaussian Process Pre-Optimization for Asynchronous Event-Inertial Odometry}
%\author{Zhixiang Wang, Yizhai Zhang,~\IEEEmembership{Member,~IEEE,} Yuchen Yan, Xudong Li, \\ Yongwei Zhang, Panfeng Huang,~\IEEEmembership{Senior Member,~IEEE}

\author{Zhixiang Wang, Xudong Li, Yizhai Zhang, Fan Zhang, and Panfeng Huang,~\IEEEmembership{Senior Member,~IEEE}
%\author{XXX

        % <-this % stops a space
\thanks{This work was supported by the National Natural Science Foundation of China under Grant Nos. 62573351 and 62022067.  (\emph{Corresponding author: Yizhai Zhang.})}% <-this % stops a space

\thanks{All authors are with Research Center for Intelligent Robotics, School of Astronautics, Northwestern Polytechnical University,  Xi'an 710072, China (e-mail: wangzhixiang@mail.nwpu.edu.cn, zhangyizhai@nwpu.edu.cn).}
}

\maketitle

\begin{abstract}
Event cameras, as bio-inspired sensors, are asynchronously triggered with high-temporal resolution compared to intensity cameras. Recent work has focused on fusing the event measurements with inertial measurements to enable ego-motion estimation in high-speed and HDR environments. However, existing methods predominantly rely on IMU preintegration designed mainly for synchronous sensors and discrete-time frameworks. In this paper, we propose GPO, a continuous-time preintegration framework that can efficiently achieve tightly-coupled fusion of fully asynchronous sensors. Concretely, we model the preintegration as two local Temporal Gaussian Process (TGP) trajectories and leverage a light-weight two-step optimization to infer the continuous preintegration pseudo-measurements. 
We show that the Jacobians of arbitrary queried states can be naturally propagated using our framework, which enables GPO to be involved in the asynchronous fusion.
Our method realizes a linear and constant time cost for optimization and query, respectively. To further validate the proposal, we leverage GPO to design an asynchronous event-inertial odometry and compare with other asynchronous fusion schemes. Experiments conducted on both public and own-collected datasets demonstrate that the proposed GPO offers significant advantages in terms of accuracy and efficiency, outperforming existing approaches in handling asynchronous sensor fusion.
The code of GPO can be found at {\url{https://github.com/NPU-RCIR/GPO-Preint}}. 
\end{abstract}

\begin{IEEEkeywords}
event-inertial fusion, Gaussian process regression, motion estimation, asynchronous fusion.
\end{IEEEkeywords}

\section{Introduction}

\IEEEPARstart{T}he estimation of ego-motion is critical for robots to accomplish automated tasks, such as bridge inspection \cite{wang2024localization}, disaster rescue \cite{albanese2021sardo}, and autonomous driving \cite{li2024occ}. Traditionally, this problem has been addressed by fusing various sensors using filter-based \cite{fornasier2023msceqf}, optimization-based \cite{zhang2022monocular}, or learning-based methods \cite{klenk2024deep}. However, these conventional discrete-time estimation frameworks struggle with handling high-frequency or even asynchronous  sensors effectively. To address this challenge, continuous-time methods have been proposed to associate measurements from different sensors with a unified, time-indexed trajectory, such as B-splines \cite{lv2023continuous} or Temporal Gaussian Processes (TGP) \cite{talbot2024continuous}.

Recent studies have demonstrated that event-inertial odometry based on continuous-time frameworks have great potential for ego-motion estimation in high-speed and High Dynamic Range (HDR) scenarios \cite{wang2024asyneio,wang2023event}. Nonetheless, existing continuous-time event-inertial systems often involve a substantial number of inertial factors and optimize them together with kinematic states of a global motion trajectory, leading to high computational consumption. Other schemes integrate inertial measurements (referred to as preintegration) between keyframes to reduce the number of inertial factors, but paying the cost of neglecting the rich motion information between integration intervals. Consequently, event factors can only rely on an imprecise prior trajectory to build re-projection relationships. Since the prior trajectory may violate the true motion, it can corrupt the estimation.

Continuous-time preintegration methods mitigate these issues by modeling the integration procedure as continuous inference using differential equations or latent states \cite{eckenhoff2020high, le2023continuous}. This enables the asynchronous fusion ability to query on a local inertial preintegration trajectory. The primary drawback is that the initialization and inference are still time-consuming for online applications. In addition. the differential equation-based methods often have less supports for asynchronous querying and back-end optimization. 
%The primary drawback of latent state preintegration is its high computational cost, which scales cubically with the size of the latent state.

To address these challenges, we propose a hybrid estimation framework that combines the advantages of both discrete- and continuous-time methods. We introduce a continuous \textbf{G}aussian \textbf{P}rocess pre-\textbf{O}ptimization (\textbf{GPO}) framework to estimate the inertial preintegration pseudo-measurements. Our GPO employs a light-weight two-step optimization and achieves linear solving and constant query times, enabling precise kinematic state queries at arbitrary time. 
%The proposed GPO is further integrated into an asynchronous event-inertial fusion system.
The asynchronous fusion ability of GPO is further validated by integrating it into an asynchronous event-inertial fusion system.   
Both the simulation and real-world experiments demonstrate the superior
efficiency and accuracy performance of our method compared to state-of-the-art ones. 
In summary, the contributions of this paper are listed as below:
\begin{enumerate}{}{}
	\item{An efficient pre-optimization method that converts raw inertial measurements to continuous local TGP trajectory, enabling constant querying time at arbitrary time. }
	
	\item{A Jacobian propagation strategy that enables adjustment of system states for any synchronous or asynchronous cost terms.  }
	
	\item{A complete asynchronous fusion pipeline for GPO, where the asynchronous fusion ability of GPO is validated by tightly-coupled fusion of event associations and inertial measurements at arbitrary timestamps.}
	
	\item{Verification and analysis of the developed system on both public datasets and real-world experiments. }
\end{enumerate}

\section{Related Work}
\label{sec:Related Work}

\subsection{Continuous-Time Estimation and Preintegration}

The concept of preintegration was initially introduced by Lupton \emph{et al.} \cite{lupton2011visual} and Forster \emph{et al.} \cite{forster2015imu}. 
Recently, Zhang \emph{et al.} \cite{zhang2024gnss} utilized the discrete-time preintegration to a global TGP-based continuous-time fusion framework. However, it can only offer a relative motion measurement between two adjacent motion states, and ignores the contribution of high-frequency motion information for asynchronous sensor fusion.
The discrete-time preintegration assumes constant measurements during each integration step, which may introduce additional errors. Differential equation-based methods \cite{shen2015tightly, eckenhoff2020high, eckenhoff2019closed} were designed to alleviate it by modeling the rigid motion using the kinematic formula. 
The Gaussian Process (GP)-based preintegration methods were proposed to model the IMU measurements as a group of latent states \cite{le2020gaussian, le20183d}. 
But it only consider the 1-axis-rotation scenario and still rely on the iterative numerical integration.
Further improvements were reported in LPM and GPM \cite{le2020in2laama, le2021continuous} to model 3-axis rotation as latent states of three independent GPs. 
They further proposed GP-Preintegration (GPP) to support both attitude and translation preintegration by six independent GPs \cite{le2023continuous}, where LPM is leveraged to finish initialization. 
Although their methods offer an asynchronous fusion capability, the high computational burden can degenerate the real-time performance.
Recently, Burnett \emph{et al.} \cite{zheng2024traj, burnett2024continuous} modeled the raw IMU measurements as the direct observations of a global TGP trajectory. However, it introduced numerous inertial factors and thus requires high computational cost.
Conversely, our proposal introduces the TGP theory into preintegration and suggests a two-step pre-optimization to instantiate a local TGP trajectory, which achieves superior efficiency.

\subsection{Event-Inertial Odometry}

Event-inertial odometry (EIO) has garnered significant attentions for its potential to estimate ego-motion in challenging scenarios \cite{gallego2020event}. 
Mahlknecht \emph{et al.} \cite{mahlknecht2022exploring} synchronized measurements from the Event-based KLT (EKLT) tracker \cite{gehrig2020eklt} via extrapolation of fixed number of events and fused them with IMU data using a filter-based back-end. 
Rebecq \emph{et al.} \cite{rebecq2017real, Vidal2018Ultimate} introduced a keyframe-based framework in which feature points from motion-compensated event frames are fused with discrete-time preintegration.
Researchers also utilized Time Surface (TS) \cite{lagorce2016hots}, which stores timestamps of most recent events at each pixel, to detect and track features using frame-based techniques \cite{guan2022monocular, tang2024monocular} and fuse them with discrete-time preintegration. 
In general, these methods transform asynchronous event streams into synchronous data associations and convert high-rate IMU data into inter-frame motion constraints through discrete-time preintegration.
However, discrete-time estimation frameworks neglect the asynchronous nature of event cameras, resulting in suboptimal performance in challenging scenarios.

Event-triggered frameworks have been proposed to preserve the high-temporal resolution and asynchronous nature of event cameras. 
Liu \emph{et al.} \cite{liu2022asynchronous} introduced a monocular asynchronous back-end (without IMU) to trigger optimization by incoming events. Wang \emph{et al.} \cite{wang2023event} developed a stereo event odometry using global TGP with an incremental back-end. Dai \emph{et al.} \cite{dai2022tightly} designed an event-inertial odometry where LPM was used to realize tightly-coupled fusion. Le \emph{et al.} \cite{le2020idol} extended this work by introducing a line feature tracker. 
Li et al. \cite{ownirosasyn} realized asynchronous fusion of event-inertial odometry using GPP. Wang \emph{et al.} \cite{wang2024asyneio} further extended this work by introducing a total asynchronous front-end and realized asynchronous fusion using various inertial schemes on global TGP. However, the initialization and query operators of the GPP are quite time-consuming. Moreover, the GPP initialized six independent GPs to fit the $SE(3)$ pose, which results in a sub-optimal attitude accuracy in practical  EIO. Instead, the proposed GPO adopts two local TGP trajectories (on $SO(3)$ and \brevised{$\mathbb{R}^{3}$}) to model the continuous-time inertial preintegration, which handles the rotation Lie group more elegantly and achieves higher accuracy and efficiency in the real EIO system. Compared with global TGP-based methods \cite{wang2024asyneio,burnett2024continuous}, our GPO excludes TGP states from the back-end optimization, thereby significantly reducing  problem dimensionality.

\section{METHODOLOGY}
\label{sec:methodology}

\begin{figure*}[!t]
	\centering
	\includegraphics[width=6.0in]{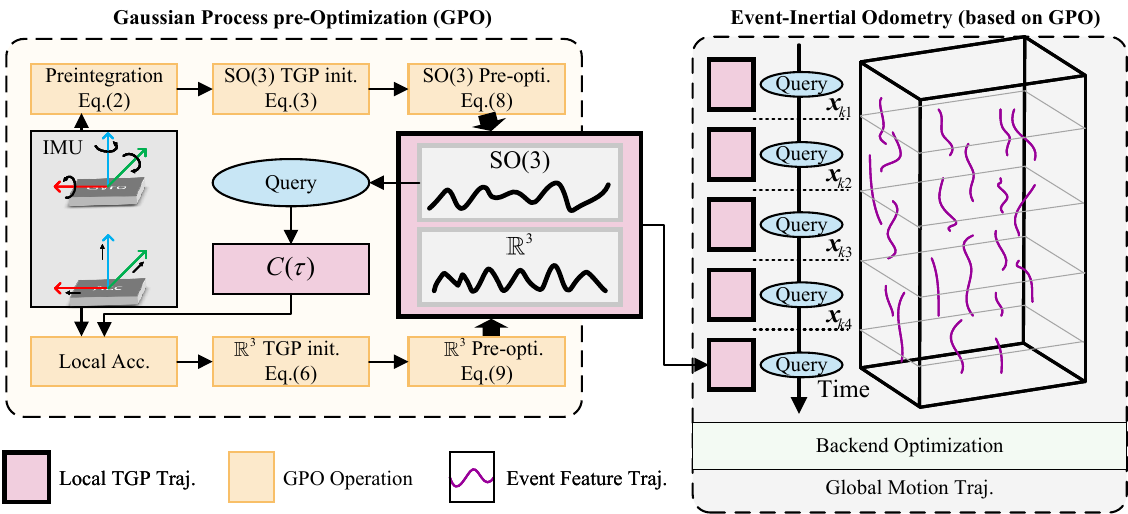}
	\caption{System framework of the proposal. The GPO infers a local TGP-based trajectory during each pre-optimization (left part). The local trajectories, within the sliding-window of event-inertial odometry, will be queried by event feature trajectories to create asynchronous projection factors (right part). }
	\label{fig:visual_abs}
\end{figure*}

\subsection{Inertial Preintegration}

Define the measurements of the gyroscope and accelerometer be $\tilde{\boldsymbol{\omega}}$ and $\tilde{\boldsymbol{a}}$, respectively. 
The IMU's measurement model is as follows:
\begin{equation}
	\label{eq:imu_measure_model}
	\begin{aligned}
		\tilde{\boldsymbol{\omega}} =\boldsymbol{\omega}_{b}^{bi} + \mathbf{b}_{g} + \boldsymbol{\varsigma}_{g}, \\
		\tilde{\boldsymbol{a}} = \boldsymbol{C}_{bi} (\ddot{\boldsymbol{r}}_{i}^{bi} -\boldsymbol{g}_{i} ) + \boldsymbol{b}_{a} + \boldsymbol{\varsigma}_{a},
	\end{aligned}	
\end{equation}
where $\boldsymbol{\varsigma}_{g} \sim \mathcal{N}(\boldsymbol{0},\boldsymbol{Q}_{g})$ and  $\boldsymbol{\varsigma}_{a}  \sim \mathcal{N}(\boldsymbol{0},\boldsymbol{Q}_{a})$ are the measurement noises, \brevised{$\boldsymbol{C}_{bi}$ the rotation matrix, $\boldsymbol{r}_{i}^{bi}$ the translation,} $\boldsymbol{g}_{i}$ the gravitational acceleration, $\mathbf{b}_{g}$ and $\mathbf{b}_{a}$ the gyroscope and accelerometer biases. 
\brevised{The observed value $\boldsymbol{\omega}_{b}^{bi}$ represents the body ($b$-superscript) angular velocity with respect to inertial navigation frame ($i$-superscript) and expressed in body frame ($b$-subscript), and the body acceleration $\ddot{\boldsymbol{r}}_{i}^{bi}$ is the second time derivative of $\boldsymbol{r}_{i}^{bi}$.}
According to previous research \cite{forster2015imu, le2023continuous},  
the inertial preintegration pseudo-measurements in the time period $[t_{n}, t_{n+1}]$ can be expressed by
\begin{equation}
	\label{eq:preinegration}
	\brevised{\begin{aligned}
		&\Delta \boldsymbol{C} = \prod_{t_{n}}^{t_{n+1}} \exp\left((\tilde{\boldsymbol{\omega}}(\tau) - \boldsymbol{b}_{g} ) d\tau\right), \\
		&\Delta \boldsymbol{v} = \int_{t_{n}}^{t_{n+1}} \boldsymbol{C}_{n}(\tau) ( \tilde{\boldsymbol{a}}(\tau) - \boldsymbol{b}_{a} ) d \tau, \\
		&\Delta \boldsymbol{r} = \int_{t_{n}}^{t_{n+1}}\!\!\!\int_{t_{n}}^{\tau} \boldsymbol{C}_{n}(s) ( \tilde{\boldsymbol{a}}(s) - \boldsymbol{b}_{a} ) ds d\tau,
	\end{aligned}}
\end{equation}
\brevised{where $\tau$ and $s$ indicate the timestamps, $\exp(): \mathbb{R}^{3} \to SO(3)$ represents the exponential map, and $\Delta \boldsymbol{C} \in SO(3), \Delta \boldsymbol{v} \in \mathbb{R}^{3}, \Delta \boldsymbol{r} \in \mathbb{R}^{3}$ is the rotation, linear velocity and translation components, respectively. As the multiplication is applied to compose on $SO(3)$, $\Delta \boldsymbol{C}$ uses the product integral. 
Note that $\boldsymbol{C}_{n}(\tau)$ is the rotation integral at time $\tau$ and can be computed from $\Delta \boldsymbol{C}$ simultaneously.} When the estimated IMU biases $\boldsymbol{b}_{a}$ and $\boldsymbol{b}_{g}$ are changed, the inertial preintegration measurements in \eqref{eq:preinegration} are updated with linearized Jacobians. 

\subsection{GP Regression on Preintegration States}
Although the inertial preintegration works well when applied in discrete-time visual-inertial system, it would be suboptimal for the asynchronous measurement fusion, such as the event-inertial odometry and incompatible frequency measurements of gyroscopes and accelerometers. We address this drawback by integrating a pre-optimization step, where raw inertial measurements are modeled as a local continuous-time trajectory  (termed \emph{pseudo-measurement trajectory})  using the sparse GP regression method. 

Given the raw measurement $\tilde{\boldsymbol{\omega}}(\tau_{i})$ and $\tilde{\boldsymbol{a}}(\tau_{j})$, $\tau_{i}, \tau_{j} \in [t_{n}, t_{n+1}]$, we expect to search the preintegration pseudo-measurements at arbitrary timestamp $\tau \in [t_{n}, t_{n+1}]$, even if the raw IMU measurements are asynchronous (i.e., $\tau_{i} \neq \tau_{j}$ when $i=j$ or $\tau_{i+1}-\tau_{i} \neq \tau_{i} - \tau_{i-1}$ or $\tau_{j+1}-\tau_{j} \neq \tau_{j} - \tau_{j-1}$). Assume the angular acceleration $\dot{\boldsymbol{\omega}}(\tau)$ to obey a zero-mean GP. The differential equation for the rotation preintegration state can be defined as
\begin{equation}
	\label{eq:rotation_preint_gaussian}
\dot{\boldsymbol{C}}(\tau) = \boldsymbol{C}(\tau) (\boldsymbol{\omega}(\tau))^{\land},  \quad
\dot{\boldsymbol{\omega}}(\tau) \sim \mathcal{GP} (\boldsymbol{0}, \boldsymbol{Q}_{c} \delta(\tau-\tau')),
\end{equation}
where $\land$ is the the skew-symmetric operator, $\delta(\tau - \tau')$ is the \emph{Dirac delta} function, and $\boldsymbol{Q}_{c}$ is the \emph{power spectral density matrix} \cite{talbot2024continuous}. 
Regarding the rotation preintegration $\Delta \boldsymbol{C}$ in \eqref{eq:preinegration} as a continuous-time process, it can be described using the GP in \eqref{eq:rotation_preint_gaussian} as 
\begin{equation}
	\label{eq:preint_meas_gp_state}
	\begin{aligned}
		\boldsymbol{C}(\tau) &= \mathbf{I}   \quad  \quad \quad\ \quad \text{if}\ \tau = t_{n}, \\
		\boldsymbol{C}(\tau) &= \Delta \boldsymbol{C} \quad \quad  \quad \text{if}\ \tau = t_{n+1}, \\
		\boldsymbol{\omega}(\tau) &=\tilde{\boldsymbol{\omega}}(\tau) - \boldsymbol{b}_{g},
	\end{aligned}
\end{equation}
where $\boldsymbol{\omega}(\tau)$ is modeled as the bias corrected angular velocity.
A local variable $\boldsymbol{\phi}_{k}(\tau)$ is introduced to convert the nonlinear differential equation \eqref{eq:rotation_preint_gaussian} to be a piece-wise linear one as
\begin{equation}
	\label{eq:local_rotation_preint_gaussian}
	\begin{aligned}
		\boldsymbol{C}_{k}(\tau) = \boldsymbol{C}_{k} \exp((\boldsymbol{\phi}_{k}(\tau))^{\land}), \\
		\ddot{\boldsymbol{\phi}}_{k}(\tau) \sim \mathcal{GP} (\boldsymbol{0}, \boldsymbol{Q}_{c} \delta(\tau-\tau')),
	\end{aligned}
\end{equation}
\brevised{where $\boldsymbol{C}_{k} = \boldsymbol{C}(\tau_{k})$ is the linearization point and $\boldsymbol{C}(\tau) \equiv \boldsymbol{C}_{k}(\tau)$ is the rotation preintegration.}
The time derivative of local variable has a relationship with the bias-corrected angular velocity as  $\boldsymbol{\omega}(\tau) = \boldsymbol{\mathcal{J}} (\boldsymbol{\phi}_k(\tau))\dot{\boldsymbol{\phi}}_k(\tau)$, where $\boldsymbol{\mathcal{J}} (\boldsymbol{\phi}_k(\tau))$ is the right Jacobian of $\boldsymbol{\phi}_k(\tau)$. 
In contrast, the state space equations for translation and velocity are  inherently linear, defined as
\begin{equation}
	\label{eq:translation_preint_gaussian}
	\dot{\boldsymbol{r}}(\tau) = \boldsymbol{v}(\tau), \quad \dot{\boldsymbol{v}}(\tau) = \boldsymbol{a}(\tau), \quad \dot{\boldsymbol{a}}(\tau) \sim \mathcal{GP}(\boldsymbol{0}, \boldsymbol{Q}_{r} \delta(\tau - \tau')),
\end{equation}
where $\boldsymbol{a}(\tau) = \boldsymbol{C}_{k}(\tau) ( \tilde{\boldsymbol{a}}(\tau) - \boldsymbol{b}_{a} )$ is a localized acceleration measurement (see Fig.~\ref{fig:visual_abs} left part) corrected by the bias, and $\dot{\boldsymbol{a}}(\tau)$ is the ``jerk'' of the preintegration trajectory. As done in \eqref{eq:preint_meas_gp_state}, we expect $\boldsymbol{r}(\tau) = \Delta \boldsymbol{r}$ and $\boldsymbol{v}(\tau) = \Delta \boldsymbol{v}$, when $\tau = t_{n+1}$. 
Both \eqref{eq:local_rotation_preint_gaussian} and \eqref{eq:translation_preint_gaussian} \brevised{have} the similar closed-form solution as
\begin{align}
	\label{eq:mean_cov_solution}
	\boldsymbol{x}_k(\tau)&=\boldsymbol{\Lambda}_{k}(\tau)\boldsymbol{x}_k(\tau_k) + \boldsymbol{\Psi}_{k}(\tau)\boldsymbol{x}_k(\tau_{k+1}), \nonumber \\
	\boldsymbol{\Lambda}_{k}(\tau)&=\boldsymbol{\Phi}(\tau,\tau_k)-\boldsymbol{\Psi}_{k}(\tau)\boldsymbol{\Phi}(\tau_{k+1},\tau_k), \nonumber \\
	\boldsymbol{\Psi}_{k}(\tau)&=\boldsymbol{Q}_{k}(\tau) \boldsymbol{\Phi}(\tau_{k+1},\tau)^{\top}\boldsymbol{Q}_{k}(\tau_{k+1})^{-1},
\end{align}
where $\boldsymbol{x}_{k}(\tau)$ is the local state variable, $\boldsymbol{\Phi}(\tau, t_{k})$ the transition function, $\boldsymbol{Q}_{k}(\tau)$ the covariance matrix, and $\boldsymbol{\Lambda}_{k}(\tau)$ and $\boldsymbol{\Psi}_{k}(\tau)$ are the interpolation coefficients derived from $\boldsymbol{\Phi}(\tau, t_{k})$ and $\boldsymbol{Q}_{k}(\tau)$.
The detailed forms of $\boldsymbol{Q}_{k}(\tau)$ and $\boldsymbol{\Phi}(\tau, \tau_{k})$ can be found in \cite{tang2019white,ownirosasyn}.
In our context, the local state variables can be defined as  $\boldsymbol{\gamma}_k(\tau) \doteq [\boldsymbol{\phi}_{k}(\tau), \dot{\boldsymbol{\phi}}_{k}(\tau)]$ for \eqref{eq:local_rotation_preint_gaussian}, and $\boldsymbol{p}_{k}(\tau) \doteq [\boldsymbol{r}(\tau), \boldsymbol{v}(\tau), \boldsymbol{a}(\tau)]$ for \eqref{eq:translation_preint_gaussian}, respectively. In fact, $\boldsymbol{p}_{k}(\tau) = \boldsymbol{p}(\tau)$ is a global state variable, we add subscript $k$ for unifying the representation.

\subsection{Two-step Pre-Optimization}
Let $\tau_{0}=t_{n}$ and $\tau_{K}=t_{n+1}$, we uniformly sample $K+1$ points within the preintegration period $[t_{n}, t_{n+1}]$. The raw IMU measurements can be instantiated as a pseudo-measurement trajectory by estimating GP states at these sampled timestamps.
Based on the foregoing pseudo-measurement model, two pre-optimization steps are adopt to infer this trajectory. Let the local TGP-based states for gyroscope preintegration be $\boldsymbol{\chi}_{k}^{g} = [\boldsymbol{C}_{k}, \boldsymbol{\omega}_{k}]$, and for accelerometer preintegration be $\boldsymbol{\chi}_{k}^{a} = [\boldsymbol{r}_{k}, \boldsymbol{v}_{k}, \boldsymbol{a}_{k}]$, where $k\in\{0,1,\cdots, K\}$.  The objective of pre-optimization, consisting of measurement residuals and GP prior residuals, can be defined as
\begin{align}
	\label{eq:gyo_residual_fun1}
	&\min_{\boldsymbol{\chi}_{k, k+1}^{g}} \sum_{i=0}^{M_{g}} \Vert \boldsymbol{e}_{g}^{i} (\tau_{k}, \tau_{k+1}) \Vert^{2}_{\boldsymbol{Q}_{g}} + \sum_{k=0}^{K-1} \Vert \boldsymbol{e}_{c}(\tau_{k}, \tau_{k+1}) \Vert^{2}_{\boldsymbol{Q}_{c}^{p}}, \\
	\label{eq:acc_residual_fun2}
	&\min_{\boldsymbol{\chi}_{k, k+1}^{a}} \sum_{j=0}^{M_{a}} \Vert \boldsymbol{e}_{a}^{j} (\tau_{k}, \tau_{k+1}) \Vert^{2}_{\boldsymbol{Q}_{a}} + \sum_{k=0}^{K-1} \Vert \boldsymbol{e}_{r}(\tau_{k}, \tau_{k+1}) \Vert^{2}_{\boldsymbol{Q}_{r}^{p}},
\end{align}
where $\boldsymbol{e}_{g}^{i}$, $\boldsymbol{e}_{a}^{j}$ indicates the related gyroscope \eqref{eq:residual_gyr} and accelerometer \eqref{eq:residual_acc} measurement residuals in $[\tau_{k}, \tau_{k+1}]$, and $\boldsymbol{e}_{c}$, $\boldsymbol{e}_{r}$ are the corresponding GP prior residuals as defined in \eqref{eq:prior_attitude} and \eqref{eq:prior_trans}, the covariance matrices $\boldsymbol{Q}_{c}^{p}$, $\boldsymbol{Q}_{r}^{p}$ can be analytical inferred from $\boldsymbol{Q}_{c}$, $\boldsymbol{Q}_{r}$ as done in \cite{tang2019white,ownirosasyn}. \brevised{These problems are solved using GTSAM \cite{wang2024asyneio}.} 
We use different subscripts $\{i,j,k\}$ to represent the possible asynchronism among raw measurements and estimated GP states.
Based on the prior assumptions in \eqref{eq:local_rotation_preint_gaussian}, \eqref{eq:translation_preint_gaussian}, the related prior residuals can be given by
\begin{align}
	\label{eq:prior_attitude}
	\boldsymbol{e}_{c}(\tau_{k}, \tau_{k+1}) &= \left[\begin{array}{cc}
		\Delta \tau_k \boldsymbol{\omega}_k - \log(\boldsymbol{C}_{k}^{-1} \boldsymbol{C}_{k+1})\\
		\boldsymbol{\omega}_k - \boldsymbol{\mathcal{J}}(\log(\boldsymbol{C}_{k}^{-1} \boldsymbol{C}_{k+1}))^{-1} \boldsymbol{\omega}_{k+1} \\
	\end{array}
	\right], \\
	\label{eq:prior_trans}
	\boldsymbol{e}_{r}(\tau_{k}, \tau_{k+1}) &= \left[\begin{array}{ccc}
		\Delta \tau_k \boldsymbol{v}_k + \frac{1}{2} \Delta \tau_{k}^{2} \boldsymbol{a}_{k}  - \boldsymbol{r}_{k,k+1}\\
		\boldsymbol{v}_k + \Delta \tau_{k} \boldsymbol{a}_{k} - \boldsymbol{v}_{k+1} \\
		\boldsymbol{a}_{k} - \boldsymbol{a}_{k+1}
	\end{array}
	\right],
\end{align}
where $\Delta \tau_{k} = \tau_{k+1} - \tau_{k}$, and $\boldsymbol{r}_{k,k+1} = \boldsymbol{r}(\tau_{k+1}) - \boldsymbol{r}(\tau_{k})$.
Since we have defined the relationship between GP states and raw IMU measurements in \eqref{eq:preint_meas_gp_state}, \eqref{eq:translation_preint_gaussian}, their corresponding residuals have concise forms as follows:
\begin{align}
	\label{eq:residual_gyr}
	\boldsymbol{e}_{g}^{i}(\tau_{k}, \tau_{k+1}) &= \tilde{\boldsymbol{\omega}}(\tau_{i})  - \tilde{\boldsymbol{b}}_{g} - \boldsymbol{\omega}_{k}(\tau_{i}), \\
	\label{eq:residual_acc}
	\boldsymbol{e}_{a}^{j}(\tau_{k}, \tau_{k+1}) &= \boldsymbol{C}_{k}(\tau_{j}) ( \tilde{\boldsymbol{a}}(\tau_{j}) -  \tilde{\boldsymbol{b}}_{a} ) - \boldsymbol{a}_{k}(\tau_{j}),
\end{align}
where $\tau_{k} \leq \tau_{i}, \tau_{j} < \tau_{k+1} $. Note that  $\boldsymbol{\omega}_{k}(\tau_{i})$ is calculated by firstly interpolating the local state $\boldsymbol{\gamma}_{k}(\tau_{i})$ using \eqref{eq:mean_cov_solution} and then $\boldsymbol{\omega}_{k}(\tau_{i}) = \boldsymbol{\mathcal{J}} (\boldsymbol{\phi}_k(\tau_{i}))\dot{\boldsymbol{\phi}}_k(\tau_{i})$. The rotation component $\boldsymbol{C}_{k}(\tau_{j})$ is interpolated using \eqref{eq:local_rotation_preint_gaussian}, \eqref{eq:mean_cov_solution}. According to the prior model, the pseudo-acceleration $\boldsymbol{a}_{k}(\tau_{j})$ on the local TGP trajectory can be obtained from $\boldsymbol{a}_{k}$ and $\boldsymbol{a}_{k+1}$ by linear interpolation. %$\boldsymbol{a}_{k}$, if $\tau_{k} \leq \tau_{j} <\tau_{k+1}$. 

\subsection{Event-Inertial Odometry} \label{sec:Event-InertialOdometry}

As shown in Fig.~\ref{fig:visual_abs}, an asynchronous event-inertial odometry is designed to validate the asynchronous fusion ability of GPO.
We adopt the same asynchronous front-end as done in \cite{wang2024asyneio}, but develop a different back-end for GPO.
In our EIO, feature trajectories are tracked from raw event streams and inherently asynchronous with arbitrary timestamps. 
We leverage the proposed two-step pre-optimization to infer a series of local pseudo-measurement trajectories online. Then, they are fixed and involved in pseudo-measurement query and Jacobian propagation for  asynchronous event projection factors. Both event projection and pseudo-measurement factors are tightly coupled in the back-end factor graph. Eventually, the expected motion states are estimated by solving the factor graph with a sliding-window optimizer. 

For a given event projection factor occurring at an arbitrary timestamp, the residual Jacobians about neighbor estimated pose and velocity can be analytically inferred with the queried pseudo preintegration measurement. The Jacobians about estimated biases need further consideration. 
The previous GPP method \cite{le2023continuous} calculates time-consuming numerical Jacobians of all latent states by extra integration steps.
Instead, the proposed GPO memorizes the procedural Jacobians on timestamps of pseudo measurements using discrete-time formula and further propagates them to the arbitrary queried time using the Chain Rule. 
Assume $\boldsymbol{e}_{ev}(\tau)$ be an asynchronous event projection residual, the Jacobians can be derived by 
\begin{equation}
	\label{eq:bias_jacobian_derivation}
	\frac{\partial \boldsymbol{e}_{ev} (\tau)}{\partial \boldsymbol{b}(\tau)} = 
	\frac{\partial \boldsymbol{e}_{ev} (\tau)}{\partial \boldsymbol{x}(\tau)} \left( \frac{\partial \boldsymbol{x}(\tau)}{\partial \boldsymbol{x}(\tau_{k})}  
	\frac{\partial \boldsymbol{x}(\tau_{k})}{\partial \boldsymbol{b}(\tau)} + \frac{\partial \boldsymbol{x}(\tau)}{\partial \boldsymbol{x}(\tau_{k+1})} 
	\frac{\partial \boldsymbol{x}(\tau_{k+1})}{\partial \boldsymbol{b}(\tau)}\right),
\end{equation}
where $\tau \in [\tau_{k}, \tau_{k+1}]$ is the projection timestamp, and $\boldsymbol{x}(\tau_{k})$ and $\boldsymbol{x}(\tau_{k+1})$ are two nearest neighboring pseudo states, and $\boldsymbol{b}(\tau)$ is the bias variable. Normally, the Jacobian $\partial \boldsymbol{e}_{ev} (\tau) / \partial \boldsymbol{x}(\tau)$ can be derived from the observation model, and the GP model of GPO can give $\frac{\partial \boldsymbol{x}(\tau)}{\partial \boldsymbol{x}(\tau_{k})}$ analytically.
The pseudo state $\boldsymbol{x}(\tau_{k})$ is obtain by composing the last estimated state $\boldsymbol{x}(t_{n})$ and later pseudo measurements $\Delta \boldsymbol{x}(\tau_{k})$,
i.e., $\boldsymbol{x}(\tau_{k}) = \boldsymbol{x}(t_{n}) \boxplus \Delta \boldsymbol{x}(\tau_{k})$, where $\boxplus$ is the plus operator for linear vectors and the matrix multiply for Lie group. Note that $\Delta \boldsymbol{x}(\tau_{k}) = \{\boldsymbol{\chi}_{k}^{g}, \boldsymbol{\chi}_{k}^{a}\}$ has been previously estimated in \eqref{eq:gyo_residual_fun1}-\eqref{eq:acc_residual_fun2}, and then used as a pseudo measurement in the backend factor graph to estimate the system state $\boldsymbol{x}(t_{n})$, as shown in Fig.~\ref{fig:visual_abs}.  Similarly, the IMU bias is \brevised{regarded} as constant within the whole integration period $[t_{n}, t_{n+1})$. Therefore, we have
\begin{equation}
	\label{eq:proceduralJacob}
	\frac{\partial \boldsymbol{x}(\tau_{k})}{\partial \boldsymbol{b}(\tau)} \approx \frac{\partial\boldsymbol{x}(\tau_{k}) }{\partial \Delta \boldsymbol{x}(\tau_{k})} \frac{\partial \Delta \boldsymbol{x}(\tau_{k})}{\partial \boldsymbol{b}(\tau_{k})},
\end{equation}
in which the procedural Jacobian  $\frac{\partial \Delta \boldsymbol{x}(\tau_{k})}{\partial \boldsymbol{b}(\tau_{k})}$ can be efficiently inferred when initializing the pseudo measurements just similarly as done in discrete-time preintegration.

\begin{figure*}[!t]
	\centering
	\includegraphics[width=6.3in]{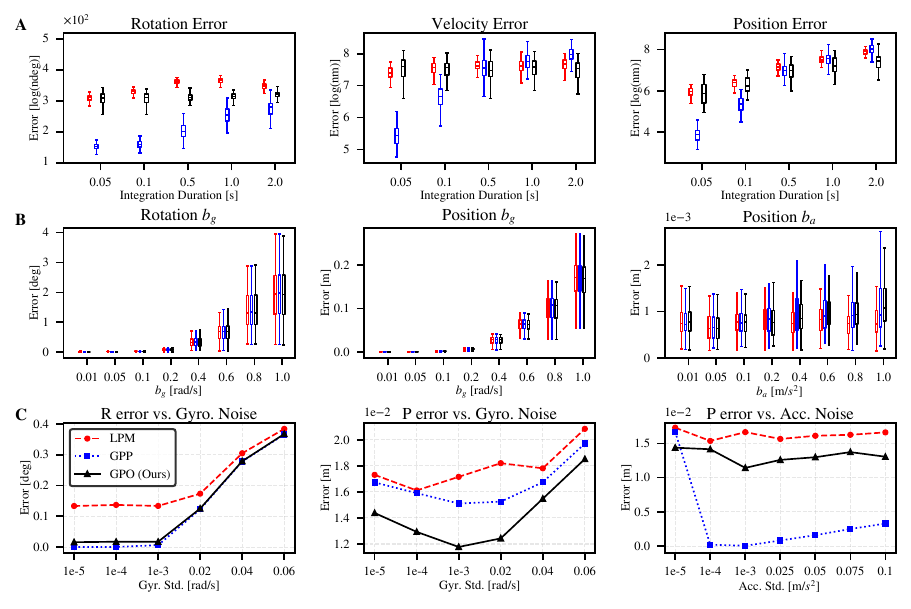}
	\caption{Simulation evaluations of the proposed GPO.  (A) Accuracy comparison under various integration durations. 
	We randomly sample 100 trials and compare their $\Delta \boldsymbol{C}$, $\Delta \boldsymbol{v}$, $\Delta \boldsymbol{r}$ with the ground truth for each integration period. (B) Procedural Jacobians evaluations. The Jacobian $\partial \Delta \boldsymbol{x}(\tau) / \partial \boldsymbol{b}(\tau)$ calculated using \eqref{eq:bias_jacobian_derivation}, \eqref{eq:proceduralJacob} is adopted to correct the integration results of known biases, and compared with the reintegration results. Ten interpolation times $\tau$ are randomly sampled within the integration duration, with 100 trials per evaluation. (C) Robustness evaluations. Raw IMU measurements are contaminated with Gaussian noise of varying magnitudes. (0.5 s, 100 Hz). }
\label{fig:gpo_eval_total_ral_revised_v2}
\end{figure*}

\section{Experiments}\label{sec:experiments}

In the experiments, we compare the proposed GPO with two similar GP-based methods, LPM \cite{le2021continuous} and GPP \cite{le2023continuous}, to evaluate their accuracy, robustness, and computational efficiency. Additionally, we evaluate the performance of our EIO system against  state-of-the-art methods by testing on public datasets and real-world experiments. 

\subsection{Precision Comparison}

\subsubsection{Preintegration Errors}
An IMU simulator \cite{le2023continuous} is developed to generate raw IMU measurements and ground truth with different motion patterns.  Then, we randomly sample the measurements and integrate them using LPM, GPP, and our GPO with five different integration periods and repeat each evaluation for 100 trials. For fair comparison, we set the same Gaussian noise ($1\times 10^{-5}\ m/s^{2}$ for accelerometer and $1\times 10^{-5}\ rad/s$ for gyroscope ) and zero bias for all six axes. The errors of preintegration are summarized in Fig.~\ref{fig:gpo_eval_total_ral_revised_v2}-A. We compare their $\Delta \boldsymbol{C}$, $\Delta \boldsymbol{v}$, $\Delta \boldsymbol{r}$ with the ground truth. The GPP achieves the highest accuracy in rotational integration across different durations. The proposed GPO consistently outperforms LPM in all scenarios, achieving superior accuracy, especially in velocity and position integration during fast motions over long durations.

\subsubsection{Jacobian Errors}
The Jacobians of preintgration results  with respect ot IMU bias $\frac{\partial \Delta \boldsymbol{x}(\tau)}{\partial \boldsymbol{b}(\tau)}$ are calculated using \eqref{eq:bias_jacobian_derivation} and \eqref{eq:proceduralJacob}. It is employed to estimate bias updates and to refine the integration results within fusion systems. Intuitively, accurate Jacobians would lead to improved estimation results. For evaluation, the ground-truth biases and the Jacobians from various methods are used to correct the preintegration results, which are then compared to the reintegration results obtained with the known biases. 
Ten interpolation times $\tau$ are randomly sampled within the integration duration, and each evaluation is conducted over 100 trials. As shown in Fig.~\ref{fig:gpo_eval_total_ral_revised_v2}-B, the proposed approximate method (see Sec. \ref{sec:Event-InertialOdometry}) achieves competitive results in all cases. 

\subsubsection{Robustness against Noise}

Inevitably, real IMU sensors contain varying degrees of noise. Therefore, robustness to noise is a crucial criterion for evaluating the performance of a preintegration method. To this end, we designed different noise levels in the simulator and calculated the preintegration errors under two motion patterns. The results are shown in Fig.~\ref{fig:gpo_eval_total_ral_revised_v2}-C. All three methods exhibit comparable sensitivity to noise for slow motion. However, when applied to fast motion, GPO and GPP demonstrate superior noise robustness. Notably, under varying gyroscope noise levels, our GPO achieves lower positional integration errors compared to GPP, indicating enhanced robustness.

\subsection{Time-Complexity Comparison} \label{time-complex-comp}
\begin{figure}[!t]
	\centering
	\includegraphics[width=3.0in]{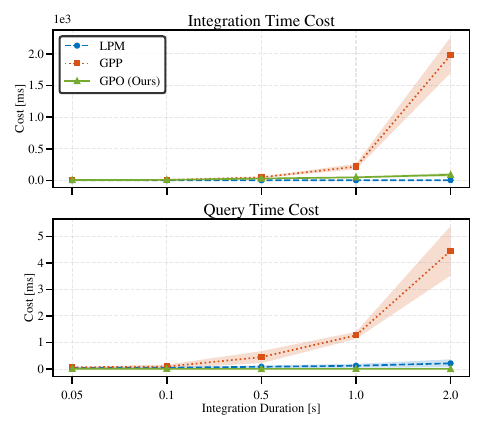}
	\caption{Time cost comparison. Top: Preintegration time - LPM is fastest (blue dashed), GPO scales linearly (green solid), GPP has cubic complexity and becomes unstable with longer durations (red dotted, with shadow showing instability). Bottom: Query time - GPO achieves constant time for pseudo-measurements, Jacobians and covariance matrices (green solid), outperforming alternatives.}
	\label{fig:time_cost_gpo}
\end{figure}

We summarize the time costs of preintegration and query, respectively. For each repeated integration trial (100 trials), 10 time points are randomly sampled within the integration duration, and the query operation is performed to compute the preintegrated quantities \{$\Delta \boldsymbol{C}(\tau)$, $\Delta \boldsymbol{v}(\tau)$, $\Delta \boldsymbol{r}(\tau)$ \} , along with the associated Jacobians and covariance. All computational efficiency evaluations are conducted on a desktop computer with an Intel Xeon Gold 6248R@4GHz, running Ubuntu 20.04 LTS. The experiments utilized only CPU resources without GPU or other parallel acceleration for the sake of fairness. The results are displayed in Fig.~\ref{fig:time_cost_gpo}.
The integration operation of the LPM resembles discrete-time preintegration and demonstrates the lowest computational cost (top, blue dashed line). Although the proposed GPO involves a two-step optimization process, it maintains high efficiency, with the computational time increasing linearly as the integration duration grows (top, green solid line). In contrast, the GPP is the most computationally expensive approach, characterized by a cubic time complexity (top, red dotted line). Furthermore, its computation time becomes increasingly unstable as the integration duration grows (top, red shaded area). Notably, our GPO achieves the best constant computational cost for querying pseudo-measurements, Jacobians, and covariance matrices (bottom, green solid line). This property makes GPO particularly well-suited for fusing high-temporal-resolution measurements from other asynchronous sensors.

\subsection{Event-Inertial Odometry Application}
Beyond preintegration evaluation, we introduce a dedicated event-inertial odometry system, where motion trajectories are inferred solely from fully asynchronous event associations and inertial measurements, in order to validate the asynchronous fusion capability of the proposed GPO.
We incorporate GPP and global TGP into the same framework for a fair comparison. The detailed description can be found in \brevised{\cite{le2023continuous, wang2024asyneio}}. 
Our event-inertial odometry system leverages a similar initialization procedure as in \cite{wang2024asyneio}. The asynchronous feature trajectories are first sampled at a fixed frequency of 20 Hz. At each sampling timestamp, we interpolate the pixel coordinates between adjacent feature points to ensure precise temporal alignment. This procedure effectively converts the asynchronous initialization problem into a synchronous one. Subsequently, the visual-inertial initialization method \cite{qin2018vins} is employed to obtain initial estimates of pose, linear velocity, and bias. Finally, the complete asynchronous feature trajectories are incorporated into a full optimization to further refine the initial states and landmarks. 
\subsubsection{Results for public datasets}
Our event-inertial odometry is compared with state-of-the-art monocular methods, including Ultimate-SLAM (USLAM) \cite{Vidal2018Ultimate}, DEVO \cite{klenk2024deep}, and PLEVIO \cite{Guan2023pl}, as well as stereo methods, including ESVIO (labeled as ESIO, E+I) \cite{chen2023esvio}, ESVIO+IMU \cite{niu2024imu}, and ESVO2 \cite{niu2025esvo2}.
To validate GPO in high-speed, aggressive, and HDR scenarios, we select DAVIS240C (DAVIS) \cite{mueggler2017event}, UZH-FPV (FPV) \cite{delmerico2019we}, VECtor \cite{gao2022vector} and HKU-Stereo (HKU-S) \cite{chen2023esvio} datasets. These datasets contain both monocular and stereo event and inertial measurements, captured by high-speed flying drones and hand-held devices with rapid shaking and rotation. The DSEC \cite{gehrig2021dsec} dataset is selected to validate the performance in large-scale environments. For fair comparisons, we align the scale of monocular methods when comparing them with stereo systems. The scale of DEVO is also corrected when compared with our event-inertial odometry.  
The Mean Position Error (MPE) and the Root Mean Square Error of rotation angles $\boldsymbol{C}_{rmse}$ are evaluated as error metrics to quantify the position and attitude errors of the estimated trajectories. For example, 
MPE=$0.88~\%$ represents the mean position error would be $0.88~m$ when the total trajectory length is $100~m$. 
The estimated results are summarized in Table~\ref{rmse_table_comp_mono_dataset} and \ref{rmse_table_comp_with_stereo_dataset}.
Our GPO achieves second best position and attitude accuracy on the DAVIS240C dataset. The high accuracy of DEVO may benefit from the precise tracking of its learning-based frontend and the use of $SIM(3)$ alignment. However, for the ultra-high-speed sequences in the UZH-FPV (labeled as FPV in Table~\ref{rmse_table_comp_mono_dataset}) dataset, the voxel grids used by DEVO can become degenerate due to motion blur. In contrast, our GPO-based pipeline achieves the best performance through high-temporal-resolution tracking and fully asynchronous fusion. USLAM (E+I) fails in most high-speed sequences, whereas PLEVIO (E+F+I) maintains its accuracy through the incorporation of intensity frames and line features. As shown in Table~\ref{rmse_table_comp_with_stereo_dataset}, our asynchronous monocular pipeline demonstrates best accuracy in most sequences of VECtor and HKU-Stereo datasets. Although the position accuracy may be facilitated by scale alignment, our method outperforms the compared stereo methods in terms of robustness under aggressive scenarios. For the large scale sequences of DSEC, the TGP-based pipeline obtains the highest precision in both attitude and position estimation. This is mainly attributed to the use of the White-Noise-on-Jerk (WNOJ) prior.  Since these sequences are captured using a car moving at nearly constant velocity with smooth rotations, the motion patterns closely align with the assumptions of the WNOJ model.  

\begin{table}[!t]
	\centering
	\renewcommand{\arraystretch}{1.1}
	\caption{The Comparative Results with Monocular Methods}
	\label{rmse_table_comp_mono_dataset}
	\resizebox{3.5in}{!}{
		\begin{threeparttable}
			\begin{tabular}{ l l cccccc}
				\toprule[1.5pt]
				\multirow{2}{*}{Dataset} & \multirow{2}{*}{Sequence} & USLAM \cite{Vidal2018Ultimate} & DEVO \cite{klenk2024deep} & PLEVIO \cite{Guan2023pl} & TGP \cite{wang2024asyneio} & GPP \cite{le2020idol} & GPO (Ours) \\
				\cmidrule(r){3-3} 		    \cmidrule(r){4-4}	           \cmidrule(r){5-5}     \cmidrule(r){6-6}  \cmidrule(r){7-7} \cmidrule(r){8-8}		  	
				& & $\boldsymbol{C}_{rmse}$  / MPE & $\boldsymbol{C}_{rmse}$  / MPE & $\boldsymbol{C}_{rmse}$  / MPE         		& $\boldsymbol{C}_{rmse}$  / MPE      	   	& $\boldsymbol{C}_{rmse}$ / MPE & $\boldsymbol{C}_{rmse}$ / MPE \\
				\cmidrule(r){1-3}	\cmidrule(r){4-4}	           \cmidrule(r){5-5}     \cmidrule(r){6-6}  \cmidrule(r){7-7}  \cmidrule(r){8-8}
				\multirow{9}{*}{DAVIS}& boxes\_6  & \underline{7.45} / 0.41 & 66.68 / 0.73 & - / \bred{0.21} & \bgreen{3.97} / 0.39 & 26.36/ 0.61 & 9.02 / \textbf{0.34} \\
				& boxes\_t   & 15.67/ 0.37 &\bgreen{3.40} / \bred{0.06} & - / \bred{0.06} & \underline{9.72} / \textbf{0.20} & 15.35/ 0.37 & 10.15/ 0.28 \\
				& poster\_6  & \bgreen{3.70}  / \textbf{0.24} & 138.64 / 0.47 & - / \bred{0.14} & \underline{5.20}  / 0.40 & 37.96/ 0.75 & 5.30 / 0.30 \\
				& poster\_t  & 6.07  / 0.25 & \underline{2.67} / \bred{0.06} & - / 0.54 & \bgreen{1.57}  / \textbf{0.14} & 7.86 / 0.26 & 4.19 / 0.21 \\
				& dynam\_6 & 5.79  / 0.34 & 4.28 / \bred{0.09} & - / 0.48 & \underline{1.13}  / 0.29 & 1.46 / 0.23 & \bgreen{0.75} / \textbf{0.11} \\
				& dynam\_t & 17.43/ 1.00 & 3.64 / \textbf{0.09} & - / 0.24 & \underline{1.13}  / 0.16 & 1.45 / 0.14 & \bgreen{0.97} / \bred{0.06} \\
				& hdr\_bo & \bgreen{2.18}  / 0.45 & 3.83 / \bred{0.07} & - / \textbf{0.10} & \underline{2.33} / 0.31 & 39.12/ 0.60 & 7.56 / 0.34 \\
				& hdr\_po& 4.65           / 0.31 & 16.41 / 0.22 & - / \bred{0.12} & \bgreen{1.53}  / 0.28 & \underline{3.30} / \textbf{0.19} & 6.83 / \textbf{0.19} \\
				\cmidrule(r){2-3}	\cmidrule(r){4-4}	           \cmidrule(r){5-5}     \cmidrule(r){6-6}  \cmidrule(r){7-7}	\cmidrule(r){8-8}
				& Avg.  & 7.87 / 0.42 & 29.94/ \bred{0.22} &-/ 0.24 & \bgreen{3.32} / 0.27 & 16.61 / 0.40 & \underline{5.60} / \textbf{0.23}  \\
				\cmidrule(r){1-3}	\cmidrule(r){4-4}	           \cmidrule(r){5-5}     \cmidrule(r){6-6}  \cmidrule(r){7-7}	\cmidrule(r){8-8}		
				\multirow{7}{*}{FPV} 
				& in\_45\_2  &\ding{55} / \ding{55}& 5.39 / 0.89 &- / -     & 3.45        / 0.91  	&\bgreen{1.88} / \textbf{0.77}   	& \underline{2.70} / \bred{0.60} \\
				&in\_45\_4 	 & - 	   / 9.79  	 &2.41  / \textbf{0.39} &- / -     &\underline{1.06} / 1.07 	&\bgreen{0.97} / 0.40	& 1.87 / \bred{0.37}  \\
				&in\_45\_9 	 & -       / 4.74     &17.60/ 1.23 &- / -     &\bgreen{3.18} / 0.92	& 4.75 		  / \textbf{0.74}	& \underline{4.24} / \bred{0.61}	\\ 
				&in\_for\_9  &\ding{55} / \ding{55}&\bgreen{1.95} / \textbf{0.53} & -/ \bred{0.44} & \underline{3.59}        / 1.56 	& 4.57        / 1.18 	& 4.13 /  1.05	\\
				&in\_for\_10 &\ding{55} / \ding{55}&\underline{2.01} / \bred{0.48} & -/ 1.06 & 2.04        / 0.91	& 3.97        / \textbf{0.63}	& \bgreen{1.75} / 0.70 \\
				\cmidrule(r){2-3}	\cmidrule(r){4-4}	           \cmidrule(r){5-5}     \cmidrule(r){6-6}  \cmidrule(r){7-7}	\cmidrule(r){8-8}
				& Avg.       &-        / 7.27     & 5.87 / \textbf{0.70}  &- / 0.75  & \bgreen{2.66}         / 1.07    &3.23         / 0.74     &  \underline{2.94}  / \bred{0.67} \\
				\bottomrule[1.5pt]
			\end{tabular}
			Units: [$\%$] for MPE, [deg] for $\boldsymbol{C}_{rmse}$. MPE is the Mean Position Error. The $SIM(3)$ alignment is applied to the estimated trajectory of DEVO, whereas the $SE(3)$ alignment is used for other event-inertial methods to ensure a fair comparison. $\boldsymbol{C}_{rmse}$ is the Root Mean Square Error of attitude matrices. 
			The bold colored numbers indicate the best precision performance among the compared methods (orange for the best MPE, green for the best $\boldsymbol{C}_{rmse}$), while underlined and bold black numbers denote the second-best precision. ``\ding{55}'' represents that the particular method fails in a sequence, and ``-'' means the corresponding value is unavailable.
			It is noteworthy that the results of all methods except PLEVIO are reproduced using their open-source codes, which may lead to discrepancies compared to the values reported in their original papers. Each evaluation was repeated more than ten times, and the best results were selected to minimize the impact of random fluctuations. 
		\end{threeparttable}
	}
\end{table}

\begin{table}[!t]
	\centering
	\renewcommand{\arraystretch}{1.1}
	\caption{The Comparative Results with Stereo Methods}
	\label{rmse_table_comp_with_stereo_dataset}
	\resizebox{3.5in}{!}{
		\begin{threeparttable}
			\begin{tabular}{ l l cccccc}
				\toprule[1.5pt]
				\multirow{2}{*}{Dataset} & \multirow{2}{*}{Sequence} & ESIO\cite{chen2023esvio} & ESVO+IMU\cite{niu2024imu} & ESVO2\cite{niu2025esvo2} & TGP \cite{wang2024asyneio} & GPP \cite{le2020idol} 	& GPO (Ours) \\
				\cmidrule(r){3-3} 		    \cmidrule(r){4-4}	           \cmidrule(r){5-5}     \cmidrule(r){6-6}  \cmidrule(r){7-7} \cmidrule(r){8-8}		  	
				& & $\boldsymbol{C}_{rmse}$  / MPE& $\boldsymbol{C}_{rmse}$  / MPE & $\boldsymbol{C}_{rmse}$  / MPE         		& $\boldsymbol{C}_{rmse}$  / MPE      	   	& $\boldsymbol{C}_{rmse}$ / MPE & $\boldsymbol{C}_{rmse}$ / MPE \\
				\cmidrule(r){1-3}	\cmidrule(r){4-4} \cmidrule(r){5-5} \cmidrule(r){6-6} \cmidrule(r){7-7} \cmidrule(r){8-8}		
				\multirow{7}{*}{DSEC} 
				&city\_04\_a & 4.83 / 4.33  &3.99  / \textbf{0.38}&\underline{3.62} / \bred{0.23}&\bgreen{1.87} / 0.58 & 3.91  / 3.59 & 6.75  / 2.65  \\
				&city\_04\_b & 10.08/ 6.02  &3.65  / 0.91 &\underline{2.90} / 1.24  &\bgreen{1.33} / 0.06 & 19.89/ \textbf{0.05}& 26.49/ \bred{0.03}	\\
				&city\_04\_c & 4.52 / 1.59  &16.29/ 1.07  &7.40 / 0.84  & 2.85 / 0.96 &\underline{2.16} / \textbf{0.47}&\bgreen{2.09} / \bred{0.22}	\\
				&city\_04\_d & 12.03/11.03 &26.12/ 1.20  &19.95/ 0.86  &\bgreen{10.13}/ \bred{0.20}&\underline{10.70}/ 0.45 & 11.03/ \textbf{0.21}	\\
				&city\_04\_e & 19.42/11.74 &13.51/ 0.82 &7.33 / 0.42  &\bgreen{1.15} / \textbf{0.27}&\underline{2.89} / \bred{0.03}& 4.26  / \bred{0.03}	\\
				&city\_04\_f & 32.62/ 9.02	&6.39    / 1.19    &12.09/ 0.50  & 7.56 / 0.47 &\bgreen{3.18} / \bred{0.11}&\underline{3.29} / \textbf{0.20}  \\
				\cmidrule(r){2-3}	\cmidrule(r){4-4}	           \cmidrule(r){5-5}     \cmidrule(r){6-6}  \cmidrule(r){7-7}	\cmidrule(r){8-8}
				& Avg.       &13.92/ 7.29   & 11.66/ 0.92  &8.88    / 0.68  &\bgreen{4.15} / \bred{0.42}  &\underline{7.12} / 0.78  &8.99 / \textbf{0.56}  \\
				\cmidrule(r){1-3}	\cmidrule(r){4-4}	           \cmidrule(r){5-5}     \cmidrule(r){6-6}  \cmidrule(r){7-7}	\cmidrule(r){8-8}
				\multirow{6}{*}{VECtor}
				& desk\_f1      &\bgreen{6.60} / 4.67 & \ding{55} / \ding{55} & \ding{55} / \ding{55} & 18.72/ 1.23 & 17.19/ \textbf{0.87}&\underline{6.74} / \bred{0.74} \\
				& hdr\_f1       & \ding{55}  / \ding{55} & \ding{55} / \ding{55} & \ding{55} / \ding{55} &\underline{4.65} / \textbf{0.72}& 6.86 / 0.92 &\bgreen{4.38} / \bred{0.62} \\
				& mount\_f1  & \ding{55}  / \ding{55} & \ding{55} / \ding{55} & \ding{55} / \ding{55} & 5.77  / \textbf{0.33}&\underline{4.94} / 0.41 &\bgreen{3.24} / \bred{0.23} \\
				& robot\_f1     & \ding{55}  / \ding{55} & \ding{55} / \ding{55} & \ding{55} / \ding{55} & 9.30  / 1.07 &\underline{3.83} / \textbf{0.62}&\bgreen{1.67} / \bred{0.42} \\
				& sofa\_f1      & \ding{55}  / \ding{55} & \ding{55} / \ding{55} & \ding{55} / \ding{55} &\underline{10.78}/ 0.73 & 11.38/ \textbf{0.68}&\bgreen{6.61} / \bred{0.45} \\
				\cmidrule(r){2-3} \cmidrule(r){4-4}	 \cmidrule(r){5-5} \cmidrule(r){6-6} \cmidrule(r){7-7}\cmidrule(r){8-8}
				& Avg.          &\underline{6.60} / 4.67 & - / - &- / - & 9.84  / 0.82 & 8.84 / \textbf{0.70} &\bgreen{4.53} / \bred{0.49}  \\
				\cmidrule(r){1-3}	\cmidrule(r){4-4}	           \cmidrule(r){5-5}     \cmidrule(r){6-6}  \cmidrule(r){7-7}	\cmidrule(r){8-8}
				\multirow{7}{*}{HKU-S}
				& agg\_flip			& 20.10/ 3.40 & \ding{55} / \ding{55} & \ding{55} / \ding{55} & 10.29/ 1.11 &\underline{9.71} / \textbf{0.81}& \bgreen{7.92} / \bred{0.64} \\
				& agg\_rot 	& 28.43/ 0.83 & \ding{55} / \ding{55} & \ding{55} / \ding{55} & 9.30  / \bred{0.16}&\underline{7.78} / \textbf{0.19}& \bgreen{1.88} / 0.23 \\
				& agg\_tran	& 17.53/ 0.57 & \ding{55} / \ding{55} & \ding{55} / \ding{55} & 7.73  / \textbf{0.22}&\underline{6.87} / \bred{0.19}& \bgreen{4.89} / \bred{0.19}\\
				& hdr\_agg      	&\underline{18.52}/ 1.44 & \ding{55} / \ding{55} & \ding{55} / \ding{55} & 22.24/ 0.60 &\bgreen{10.68}/ \bred{0.48}& 19.46/ \textbf{0.55}\\
				& hdr\_circle   	& 13.95/ 0.50 & \ding{55} / \ding{55} & \ding{55} / \ding{55} & 7.94  / \textbf{0.49}&\underline{6.93} / 0.85 &\bgreen{6.01} / \bred{0.32}\\
				& hdr\_tran\_rot	& 23.85/ 0.95 & \ding{55} / \ding{55} & \ding{55} / \ding{55} &\bgreen{1.84} / \bred{0.30}&\underline{6.65} / 0.45 & 7.35 / \textbf{0.41} \\
				\cmidrule(r){2-3} \cmidrule(r){4-4}	 \cmidrule(r){5-5} \cmidrule(r){6-6} \cmidrule(r){7-7}\cmidrule(r){8-8}
				& Avg.          	& 20.40/ 1.28 & - / - & - / -    & 9.89  / \textbf{0.48}&\underline{8.10} / 0.50 &\bgreen{7.92} / \bred{0.39}  \\
				\bottomrule[1.5pt]
			\end{tabular}
			Units: [$\%$] for MPE, [deg] for $\boldsymbol{C}_{rmse}$. MPE is the Mean Position Error where the $SE(3)$ transformation is used to align the estimated trajectory and ground truth for stereo methods, while the $SIM(3)$ transformation is leveraged to align estimated trajectories of monocular methods. The results for ESIO, ESVO+IMU, and ESVO2 on the DSEC dataset are obtained using their public available raw trajectories, while other unavailable results are reproduced using their source codes with recommended parameters and configurations.
		\end{threeparttable}
	}
\end{table}

\begin{table}[!t]
	\centering
	\renewcommand{\arraystretch}{1.1}
	\caption{The RMSE of Own-collected data}
	\label{rmse_table_own_dataset}
	\resizebox{3.0in}{!}{
		\begin{threeparttable}
			\begin{tabular}{ l rlrlrlrl}
				\toprule[1.5pt]
				\multirow{2}{*}{Sequence} & \multicolumn{2}{c}{USLAM \cite{Vidal2018Ultimate}} & \multicolumn{2}{c}{TGP \cite{wang2024asyneio}}  & \multicolumn{2}{c}{GPP \cite{le2020idol}} 	& \multicolumn{2}{c}{GPO (Ours)} \\
				\cmidrule(r){2-3} 		    \cmidrule(r){4-5}	           \cmidrule(r){6-7} \cmidrule(r){8-9}
				& $\boldsymbol{C}_{rmse}$  & MPE & $\boldsymbol{C}_{rmse}$  & MPE         		& $\boldsymbol{C}_{rmse}$  & MPE      	   	& $\boldsymbol{C}_{rmse}$ & MPE  \\
				\cmidrule(r){1-3}									   \cmidrule(r){4-5}	        \cmidrule(r){6-7} \cmidrule(r){8-9}
				indoor\_02 			  	&7.16&2.83&6.30& 1.15  			& 6.51 & 1.38   			& \bgreen{2.90} & \bred{0.86} \\
				indoor\_03 	 	      	&98.18&5.47&4.08  & \bred{0.85} 		& 5.07 & 2.03			& \bgreen{2.32} & 1.02  \\
				outdoor\_01 		  	&9.87&4.50&2.66  & 2.30			& 0.73 & 2.47				& \bgreen{0.53} & \bred{2.15}	\\
				outdoor\_02       		&\ding{55}&\ding{55}&1.64  & \bred{0.37} 	& 1.47 & 1.05 			& \bgreen{0.74} & 0.92	\\
				outdoor\_03       		&\ding{55}&\ding{55}&1.03  & 0.45			& 0.61 & \bred{0.15}		& \bgreen{0.49} & 0.15 \\
				outdoor\_05       		&11.11&2.92&3.37  & 1.80 		& 6.44 & 2.37				& \bgreen{1.60} & \bred{1.12}	\\
				outdoor\_06       		&17.60&8.36&\bgreen{1.50}&\bred{0.32} & 3.28 & 0.46		& 1.64 & 0.54	\\
				\cmidrule(r){1-3}								\cmidrule(r){4-5}	        \cmidrule(r){6-7}  \cmidrule(r){8-9}
				Avg.    			    & - & - &2.94	& 1.03	& 3.44 & 1.42 			& \bgreen{1.46} & \bred{0.97} \\
				\bottomrule[1.5pt]
			\end{tabular}
			Units: [$\%m$] for MPE, [deg] for $\boldsymbol{C}_{rmse}$. All methods are aligned with the ground truth using the position and yaw angle transformation.
		\end{threeparttable}
	}
\end{table} 

\subsubsection{Results for own-collected data}
\begin{figure}[!t]
	\centering
	\includegraphics[width=2.7in]{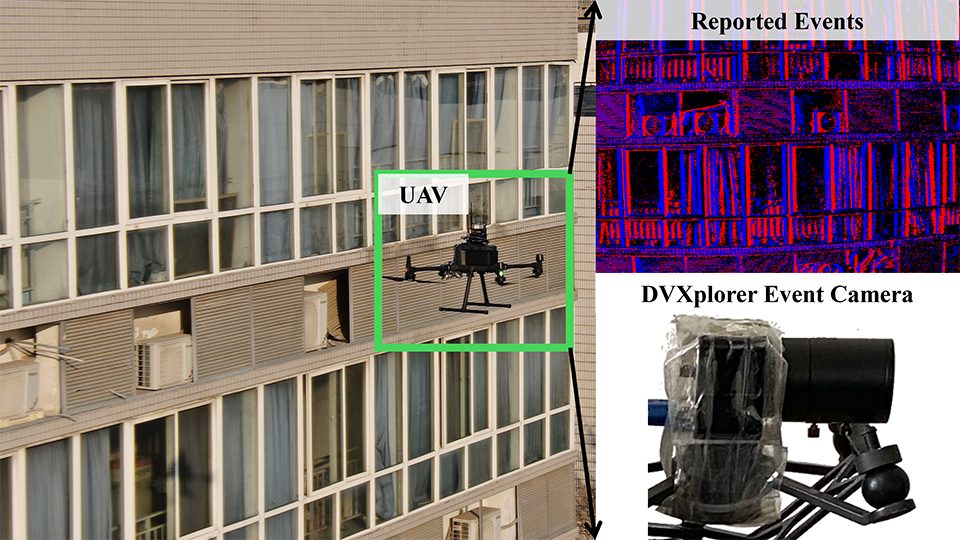}
	\caption{Real-world experiment scenario.}
	\label{fig:experiment_scenario}
\end{figure}

\begin{figure}[!htbp]
	\centering
	\includegraphics[width=3.3in]{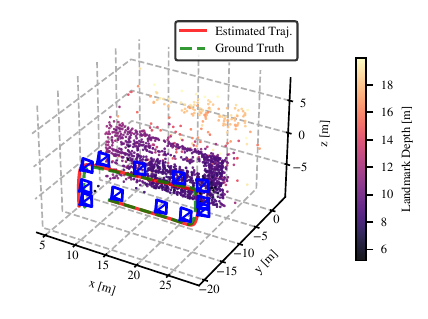}
	\caption{Estimated trajectory and landmarks for outdoor\_05.}
	\label{fig:3d_pointcloud_pose}
\end{figure}

The DVXplorer ($640 \times 480$ pixels) event camera is mounted to a DJI M300 quadrotor (see Fig.~\ref{fig:experiment_scenario}). The event streams of camera and the inertial measurements of the quadrotor are simultaneously collected together with ground truth. Ground truth is provided by an Optitrack motion capture system for indoor sequences and by Real-Time Kinematic (RTK) for outdoor scenarios. 
The quadrotor mainly executes aggressive translational motion, which means the baseline trajectories have smaller variations in yaw angles, as displayed in Fig.~\ref{fig:3d_pointcloud_pose}. 
The roll and pitch variations are mainly caused by the acceleration of the drone.
As shown in Table~\ref{rmse_table_own_dataset}, our GPO-based method outperforms the others in both attitude and position estimation. 
Compared with global TGP, GPO improves the position and attitude accuracy by $5.83 \%$ and $50.34 \%$. 
Intuitively, the GPO can be regarded as a low-pass filter for the noisy inertial measurements, thereby it can reduce the accidental measurement errors in raw measurements. 
Otherwise, its continuous-time integration model can normally achieve higher accuracy than discrete-time preintegration.
In summary, the GPO will become a more efficient and precise candidate for fusing totally asynchronous measurements with IMU.


\subsubsection{Efficiency Analysis of the Odometry System}
To assess the time efficiency of the GPO method in practical odometry systems, the computational breakdown for each module is recorded and compared with the GPP-based odometry. Since the asynchronous front-end processes raw event streams in an event-by-event manner, the average processing time per event, rather than per frame, is reported, as shown in Table~\ref{table:time_cost}. Despite the time cost being reduced to $\mu s$ magnitude, the serial processing still introduces a time lag that is roughly two times the real-time requirement. As our GPO obtains constant time cost for querying pseudo-states, the GPO-based method achieves superior time efficiency for constructing factor graph and back-end optimization, as in Table~\ref{table:time_cost}. The higher integration efficiency of GPP primarily results from the segmented parallel solving strategy during testing. A similar approach could be adopted for GPO in the future to further enhance its integration performance.

\begin{table}[!t]
	\centering
	\renewcommand{\arraystretch}{1.1}
	\caption{The Runtime Comparison of Each Module}
	\label{table:time_cost}
	\resizebox{2.5in}{!}{
		\begin{threeparttable}
			\begin{tabular}{ l ccccc}
				\toprule[1.5pt]
				Sequence & Method & Front-End & Preintegration & Construct Factors & Optimization \\
				\cmidrule(r){1-6} 
				\multirow{2}{*}{dynam\_6} & GPP & \multirow{2}{*}{$3.17 \times 10^{-3}$} & \textbf{7.44} & 21.54 & 826.52 \\
				&GPO & & 14.33 & \textbf{13.68} & \textbf{461.15} \\
				\cmidrule(r){1-6} 
				\multirow{2}{*}{desk\_f1} & GPP & \multirow{2}{*}{$5.82 \times 10^{-3}$} & \textbf{0.89}  & 7.70 & 292.81 \\
				&GPO & & 1.73 & \textbf{7.37} & \textbf{240.52} \\
				\cmidrule(r){1-6} 
				\multirow{2}{*}{hdr\_circle} & GPP & \multirow{2}{*}{$4.01 \times 10^{-3}$} & \textbf{7.33} & 12.76 & 452.48  \\
				&GPO & & 11.30 & \textbf{5.87} & \textbf{375.85} \\
				\bottomrule[1.5pt]
			\end{tabular}
			Units: [$ms/event$] for the asynchronous Front-end, $[ms]$ for other modules. Bold numbers indicate higher time efficiency.
		\end{threeparttable}
	}
\end{table} 

\section{Conclusion}
\label{sec:conclusion}

In this study, we proposed GPO, an efficient local TGP-based preintegration method specially designed for asynchronous fusion. We evaluated its performance against other GP-based methods, LPM, GPP, and global TGP. The experimental results demonstrate that GPO effectively balances accuracy, robustness, and computational efficiency.  
GPO achieves a constant time cost for querying pseudo-measurements, Jacobians, and covariance matrices, and linear cost for preintegration. The local pre-optimization can reduce the complexity of back-end optimization and improve the estimation accuracy. Therefore, we suggest the GPO be a new optional scheme for asynchronous sensor fusion.

\bibliographystyle{IEEEtran}
\bibliography{IEEEabrv,mybibfile}

\end{document}